\documentclass[conference]{IEEEtran}
\IEEEoverridecommandlockouts
\usepackage{cite}
\usepackage{amsmath,amssymb,amsfonts}
\usepackage{textcomp}
\usepackage{xcolor}

\usepackage{booktabs}

\usepackage{graphicx}
\usepackage[numbers]{natbib}
\usepackage{caption}
\usepackage{multirow}
\usepackage{algorithm}
\usepackage{algorithmic}
\usepackage{newfloat}
\usepackage{listings}
\usepackage{enumitem}
\floatname{listing}{Listing}

\newtheorem{mydef}{Definition}
\def\BibTeX{{\rm B\kern-.05em{\sc i\kern-.025em b}\kern-.08em
    T\kern-.1667em\lower.7ex\hbox{E}\kern-.125emX}}

\title{A Probabilistic Framework for Adapting to Changing and Recurring Concepts in Data Streams}
\author{\IEEEauthorblockN{Ben Halstead\IEEEauthorrefmark{1},
Yun Sing Koh\IEEEauthorrefmark{1},
Patricia Riddle\IEEEauthorrefmark{1}, 
Mykola Pechenizkiy\IEEEauthorrefmark{2},
Albert Bifet\IEEEauthorrefmark{3}}
\IEEEauthorblockA{\IEEEauthorrefmark{1}School of Computer Science, University of Auckland, New Zealand\\
Email: bhal636@aucklanduni.ac.nz, ykoh@cs.auckland.ac.nz, pat@cs.auckland.ac.nz}
\IEEEauthorblockA{\IEEEauthorrefmark{2}Eindhoven University of Technology, The Netherlands, Email: m.pechenizkiy@tue.nl}
\IEEEauthorblockA{\IEEEauthorrefmark{3}University of Waikato, Hamilton, New Zealand, and  LTCI, T{\'e}l{\'e}com Paris, IP-Paris, Email: abifet@waikato.ac.nz}}

\begin{document}
\maketitle

\begin{abstract}
The distribution of streaming data often changes over time as conditions change, a phenomenon known as concept drift.
Only a subset of previous experience, collected in similar conditions, is relevant to learning an accurate classifier for current data. Learning from irrelevant experience describing a different concept can degrade performance.
A system learning from streaming data must identify which recent experience is irrelevant when conditions change and which past experience is relevant when concepts reoccur, \textit{e.g.,} when weather events or financial patterns repeat.
Existing streaming approaches either do not consider experience to change in relevance over time and thus cannot handle concept drift, or only consider the recency of experience and thus cannot handle recurring concepts, or only sparsely evaluate relevance and thus fail when concept drift is missed.
To enable learning in changing conditions, we propose SELeCT, a probabilistic method for continuously evaluating the relevance of past experience.
SELeCT maintains a distinct internal state for each concept, representing relevant experience with a unique classifier.
We propose a Bayesian algorithm for estimating state relevance, combining the likelihood of drawing recent observations from a given state with a transition pattern prior based on the system's current state. 
The current state is continuously maintained using a Hoeffding bound based algorithm, which unlike existing methods, guarantees that every observation is classified using the state estimated as the most relevant, while also maintaining temporal stability.
We find SELeCT is able to choose experience relevant to ground truth concepts with recall and precision above 0.9, significantly outperforming existing methods and close to a theoretical optimum,  leading to significantly higher accuracy and enabling new opportunities for learning in complex changing conditions.
\end{abstract}


\section{Introduction}
\begin{figure}
    \centering
    \includegraphics[width=0.475\textwidth]{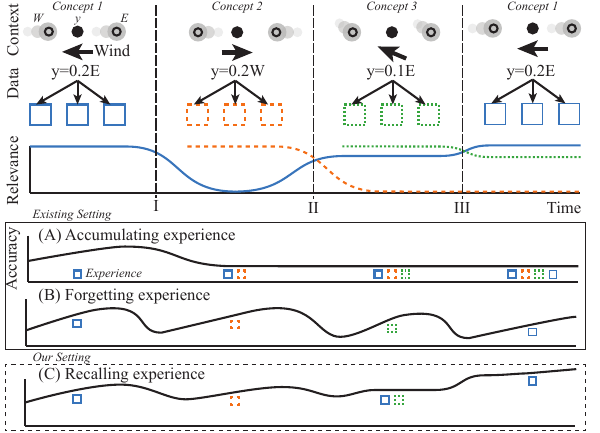}
    \caption{Relevance of experience over concept drift, and the effect on accuracy over time when accumulating, forgetting and additionally recalling experience.}
    \label{fig:intro-case-study}
\end{figure}
\begin{figure*}
    \centering
    \includegraphics[width=0.9\textwidth]{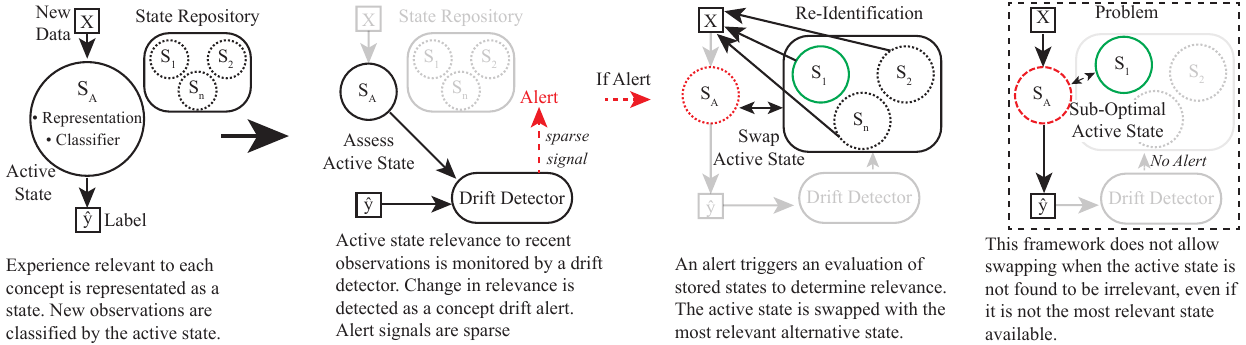}
    \caption{Standard Adaptive Learning Framework}
    \label{fig:BaselineDiag}
\end{figure*}
A growing number of systems collect data in real-time, such as internet-connected sensors or online activity.
Learning from these data streams requires adaptive systems capable of reacting to change.
Consider the data stream classification task shown in Figure~\ref{fig:intro-case-study} where the air quality level at $y$ is forecast from air pollution readings captured at $W$ and $E$. New observations may arrive at a fast pace in volumes that may not fit in memory, requiring the distribution of data, the \textit{concept}, to be learned incrementally by accumulating experience over time~\cite{gama2003accurate}.
The distribution of data is determined by the \textit{context}~\cite{gama2014recurrent}, unknown features which effect classification.
For example, wind direction is an unknown factor determining how pollution affects air quality. 
While the wind direction is \textit{constant} a classifier may accumulate experience to learn a forecasting function which is accurate for the current concept.
However, in a streaming setting context may change over time~\cite{gama2014recurrent}, causing \textit{concept drift}~\cite{tsymbal2004problem} where the distribution of incoming data changes. 
Experience learned under one concept may not generalize to other concepts, \textit{e.g.,} if wind direction changes then the forecasting function learned by the current classifier may \textit{conflict} with the new distribution of data~\cite{kessler2021same}. The task shown in Figure~\ref{fig:intro-case-study} has three distinct wind directions, with the east and west directions producing conflicting concepts. The accuracy of a system which only accumulates experience will degrade when concept drift occurs at point (I).
Experience made irrelevant due to concept drift must be \textit{forgotten} in order to maintain accuracy.
Most existing methods of learning from streaming data focus on the two challenges of accumulating experience and forgetting irrelevant experience~\cite{parisi2019continual}.


 
 In this work we investigate the additional requirement of \textit{recalling} forgotten experience when concepts reoccur, \textit{e.g.,} when similar wind conditions appear in multiple discrete periods across a stream.
Recalling relevant experience from previous occurrences of a concept is required for long-term accumulation of experience in the presence of concept drift, particularly for handling rare or short-term concepts where we cannot learn an accurate predictive function over one occurrence.
In these cases recalling experience is difficult as it requires the conditions where experience is relevant to be accurately represented, however building such a representation across concept drift requires accurately recalling experience in the first place.
Solving these dual problems requires a temporally stable system, capable of learning a stable initial representation of each concept.
Learning from streaming data with changing and recurring conditions is an open research challenge, requiring \textit{accumulating} experience over time, \textit{forgetting} experience made irrelevant by concept drift, and \textit{recalling} previous experience made relevant when concepts reoccur.
Existing streaming approaches fall into three broad categories, which all face challenges in this setting.
Approaches that do not consider the relevance of experience, or do not consider changes in relevance over time, (A) in Figure~\ref{fig:intro-case-study},  cannot adapt to concept drift and are not suitable in changing conditions. 
Approaches that determine relevance based only on forgetting irrelevant experience, (B) in Figure~\ref{fig:intro-case-study}, encounter \textit{catastrophic forgetting} where forgotten experience becomes relevant in the future, hindering the learning of long-term or recurring concepts. 
The final category, \textit{adaptive learning} methods, monitor recent observations to determine changes in relevance over time to recall recurring concepts. However, in this work we identify a flaw in existing approaches in that the relevance of past experience is only evaluated \textit{sparsely} when current experience is deemed irrelevant. Existing methods cannot guarantee that each prediction is made using relevant experience, failing at point (III) in Figure~\ref{fig:intro-case-study} where the most relevant experience cannot be recalled because current experience is not considered irrelevant.

In this work, we address these challenges by proposing a framework for adapting to concept drift able to guarantee that each prediction uses the experience estimated as most relevant, producing (C) in Figure~\ref{fig:intro-case-study}.
Similarly to  adaptive learning, our approach accumulates experience relevant to each concept into a repository of internal \textit{states} each represented by a distinct classifier.
By selecting a relevant \textit{active} state from the repository to classify and learn from each incoming observation, experience can be forgotten and recalled dynamically in response to changing distributions.

In existing methods, the relevance of the active state is monitored by a \textit{concept drift detector}~\cite{gama2004learning, barros2018large}, producing an alert when changes in distribution occur, deactivating the active state to forget irrelevant experience and selecting a new active state to handle the new concept. 
A re-identification~\cite{halstead2021fingerprinting} procedure queries previously deactivated states to determine if any are relevant to the new concept, or, if no suitable state is found, the concept is deemed novel and a new active state is initialized.
In this procedure, the relevance of the active state is continuously monitored, identifying when relevant experience becomes irrelevant, however the relevance of inactive states is only \textit{sparsely} evaluated when drift is detected, which does not identify drift where irrelevant experience becomes relevant, as shown in Figure~\ref{fig:BaselineDiag}.
Additionally, relevance in existing methods is often a \textit{binary} signal due to separating detection and re-identification. We often cannot compare \textit{how} relevant experiences are, causing failure cases when concept drift detection and re-identification conflict.

We propose an alternative approach to determining state relevance, \textbf{S}treaming \textbf{E}xtrac\-tion of \textbf{L}ik\textbf{e}lihoods for \textbf{C}oncept \textbf{T}ransitions, (SELeCT), which integrates concept drift detection and re-identification into a single, probabilistic algorithm able to continuously evaluate how relevant each state is to current observations. We propose a Bayesian algorithm to compute the relevance of a state by combining the likelihood of drawing recent observations with a prior probability learned from previous concept transitions. 
To maintain the temporal stability of states, an important property for learning each distinct underlying concept, we propose a continuous selection algorithm based on the Hoeffding bound, giving a guarantee within some risk level that the selected state at any point in the stream has the highest estimated relevance to the underlying concept displayed in recent observations, even in noisy conditions.
Our evaluation shows that the SELeCT can choose states that match ground truth concepts with a recall and precision above 0.9, showing no significant difference compared to a theoretically optimal selection strategy and significantly outperforming existing methods. 
By more accurately selecting relevant experience, we show that SELeCT is able to achieve a classification $\kappa$ statistic up to 0.15 higher than existing methods.
Our contributions are as follows:
\begin{itemize}[topsep=0pt]
    \item Our main contribution is proposing the first method of explicitly accumulating, forgetting and recalling experience relevant to changing conditions in a continuous, probabilistic manner, enabling us to learn recurring concepts.
    
    \item We develop: 1) a method of estimating relevance based on the likelihood of drawing recent observations and a prior probability encoding learned transition patterns. 2) A continuous selection algorithm based on the Hoeffding bound, able to guarantee, even in noisy real-world conditions, that the experience used for prediction is the best choice given our estimator, and, crucially, is temporally stable, enabling us to learn a representation of each concept. 3) A strategy for merging states with similar relevance dynamics to reduce memory costs.
    
    
    
    \item We contribute a novel adaptive learning framework, SELeCT, enabling near-optimal adaptations to changing conditions. 
    We evaluate SELeCT in streaming tasks with recurring concepts using decision tree and neural network classifiers, demonstrating accuracy up to 43\% higher than existing adaptive learning methods.
    
\end{itemize}

\section{Problem Formulation}\label{sec:background}
\begin{figure*}[t]
    \centering
    \includegraphics[width=0.9\textwidth]{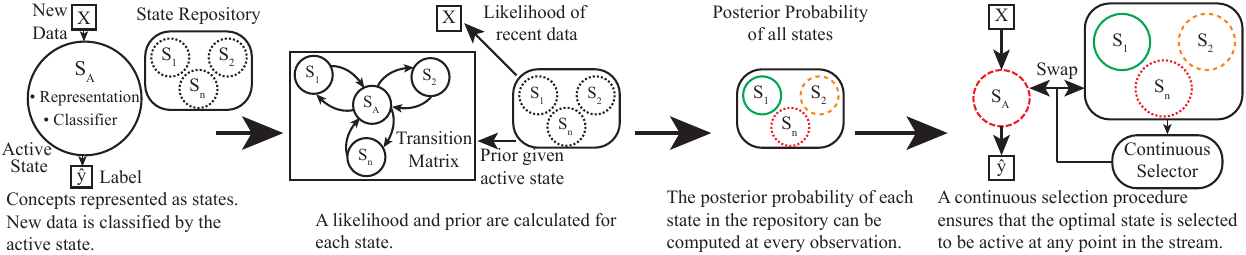}
    \caption{SELeCT Framework}
    \label{fig:SELECTDiag}
\end{figure*}
We consider the supervised data stream classification problem, where a sequence of $\langle X, y \rangle$ observations is received over time. We consider labels to be immediately available, however our framework generalizes to the unsupervised setting if an unsupervised concept representation~\cite{halstead2021fingerprinting} is used.
Each observation at time $t$, $O^t$, occurs in some \textit{context}, $H^t$, the set of all unobserved features which affect $y$~\cite{widmer1996learning}.
Each observation is drawn from the distribution $p(X, y | H^t)$, known as the \textit{concept} at time $t$.
As $H^t$ is unobserved, we denote the concept at $t$ as $C^t = p_t(X, y)$.

It is common for $H^t$ to change over time, leading to observations at different points in a data stream being drawn from different concepts. A \textit{concept drift} is a timestep $t_D$ where $p_{t<t_D}(X, y) \ne p_{t>t_D}(X, y)$~\cite{gama2014survey}.
A concept drift can be gradual, occurring over a period of observations, or abrupt~\cite{webb2016characterizing}. 
Assuming that context changes smoothly over time, concepts are \textit{temporally stable}, \textit{i.e.,} a data stream can be viewed as a sequence of segments, each containing observations drawn from a distinct concept, separated by concept drift. Concepts can reoccur over the stream, \textit{i.e.,} distinct segments may share a distribution.
Classification in this setting is difficult, as experience describing one distribution may not accurately describe a different distribution. A classifier trained on experience from one concept may inaccurately predict observations drawn from another. We define experience, or a classifier trained on this experience, as \textit{relevant} to a concept if it accurately describes the distribution of data drawn from that concept or \textit{irrelevant} if it does not. To maximize accuracy we aim to learn each prediction  from all \textit{relevant} previous experience without \textit{irrelevant} experience.

{\bf Adaptive learning} approaches accumulate new experience as distinct \textit{states}, $S_i = \langle m_i, \zeta_i \rangle$, defined in Definition~\ref{def:state}, each representing a specific concept $C_i$.
At time $t$, we have access to a \textit{repository} $R$ of previously constructed states, and we can create a new \textit{background} state $B$ from a window of recent observations $\omega$.
The goal of adaptive learning is to select the \textit{optimal} state from $R \cup B$ to classify each observation. 
Definition~\ref{def:optimal-state} defines optimal where $\zeta_i$ fully describes $C_i$ and $sim$ is a perfect measure of distribution similarity. In the real-world $\zeta_i$ is an approximation so our similarity measures are estimates, $sim^{\prime}$, thus we find the \textit{optimal estimated state}.
\begin{mydef}\label{def:state}
A \textbf{state} $S_i = \langle m_i, \zeta_i \rangle$ represents a concept $C_i$ as a relevant classifier $m_i$, \textit{i.e.,} trained on observations drawn from $C_i$, and a \textit{concept representation}, $\zeta_i$, summarizing the distribution $C_i$.
\end{mydef}

\begin{mydef}\label{def:optimal-state}
Given current concept $C_i$, repository $R$, background state $B$, a distribution similarity measure $sim$, estimated as $sim^{\prime}$ we define the
\textbf{Optimal state} as $S_o = \arg \max_{S_j \in \{R \cup B\}} sim(\zeta_j, C_i)$ and the
\textbf{Optimal Estimated state} as $S^{\prime}_o = \arg \max_{S_j \in \{R \cup B\}} sim^{\prime}(\zeta_j, C_i)$
\end{mydef}
An adaptive learning approach aims to, at each $t$, select the state $S_A^t$ which best approximates the optimal state under a given $sim^{\prime}$, $S_A^t = S^{\prime t}_o \approx S^t_o$.
Identifying the optimal state for each observation achieves our problem formulation, assuming that experience is relevant to a specific concept.  While learning, selecting the optimal state allows us to accumulate all experience relevant to each concept into a single state, containing no irrelevant experience, which we can recall to optimally predict future observations.
Additionally, selecting optimal states allows the dynamics of $S_A^t$ to be mined to reveal the dynamics of the hidden context $H^t$ to, for example, reveal important features not currently included in $X$~\cite{borchani2015modeling}.


Existing adaptive learning methods, discussed in Section~\ref{sec:related-work}, select $S_A^t$ by following the procedure shown in Figure~\ref{fig:BaselineDiag}. Initially, an \textit{active} state $S_A$ is constructed to learn from new observations. A concept drift detector monitors the relevance of the active state, producing an alert when significant change is detected in a similarity measure $sim^{\prime}(\zeta_A, \zeta_\omega)$, where $\zeta_\omega$ describes a window $\omega$ of recent observations. 
An alert indicates $S_A$ is no longer relevant, \textit{i.e.,} should be deactivated and stored in $R$, and a new $S_A$ should be selected.
A re-identification procedure can query $R$ for previous $S_i \in R$ which have become relevant, \textit{i.e.,} show a high $sim^{\prime}(\zeta_i, \zeta_\omega)$. If no $S_i \in R$ is relevant, we determine the new concept is novel and initialize a new $S_A$. 
This approach only evaluates the relevance of $S_i \in R$ \textit{sparsely}, when $S_A$ is found to be irrelevant and an alert is triggered.
While efficient, this approach cannot identify concept drift which presents as an increase in the relevance of a state in $R$, rather than a significant decrease in the relevance of $S_A$, \textit{e.g.,} due to noise or partial changes.
Sparse evaluation cannot be solved by simply running re-identification every step, as re-identification does not maintain a temporally stable active state, \textit{i.e.,} $S_A$ may change at any $t$ due to noise. This behaviour does not reflect true change in context, and means we cannot accumulate experience of each distinct concept. We require an efficient method of continuously evaluating relevance that maintains a temporally stable active state.


In Section~\ref{sec:Evaluation}, we  empirically show that the active state in existing methods often fails to capture the ground truth concept, causing accuracy to be degraded.
Alternative stream learning approaches, such as dynamic selection and continual learning discussed in Section~\ref{sec:related-work}, cannot explicitly detect irrelevant experience, or store and recall relevant past experience.
In the next section we propose SELeCT to accurately identify relevant experience in streaming data with changing and recurring concepts and show that we achieve classification performance and context tracking closer to optimal.
\begin{algorithm}[tb]
\caption{SELeCT}
\label{alg:algorithm}
\textbf{In}: New observation $O^t$ (Input $X^t$, Class $y^t$), State repository $R$, Background state $B$, Active state $S_A^t$\\
\textbf{Out}: Label $\hat{y}^t$, Next state $S_A^{t+1}$
\begin{algorithmic}[1] 
\STATE $\hat{y}^t\leftarrow Classify(S_A^t, X^t)$.
\STATE $Train(S_A^t, y^t)$
\STATE $StateProbabilities \leftarrow []$
\FOR{$S_i \in R\cup B$}
\STATE $K^t \leftarrow CurrentKnowledge(S_A^t, O^t)$
\STATE $p_i \leftarrow ComputePrior(K^t, S_i)$
\STATE $l_i \leftarrow UpdateLikelihood(O^t, S_i)$
\STATE $StateProbabilities[S_i] \leftarrow p_i \times l_i$
\ENDFOR
\STATE $S_A^{t+1} \leftarrow SelectionTest(StateProbabilities, S_A^t)$
\IF{$S_A^{t+1} = B$}
\STATE $R \leftarrow R \cup B$, $B \leftarrow NewState()$
\ENDIF
\STATE \textbf{return} $\hat{y}^t, S_A^{t+1}$
\end{algorithmic}
\end{algorithm}
\section {SELeCT Framework Overview}\label{sec:SELECT}
SELeCT is an adaptive learning framework based around collecting a repository of states, each accumulating experience relevant to a distinct concept, and identifying the optimal state to handle each observation. The main novelty is that, rather than a cycle of continuous concept drift detection and sparse re-identification, SELeCT continuously evaluates the relevance of all states, enabling us to detect both when the active state becomes irrelevant and when an inactive state becomes relevant. A continuous selection method using the Hoeffding bound solves the temporal stability challenge, guaranteeing that every active state $S_A^t$ is the optimal estimated state $S^{\prime t}_o$ on recent data, even in noisy conditions. By maintaining temporally stable active states we match the dynamics of the underlying concepts and can learn experience relevant to each distinct concept.
SELeCT is an abstract framework, so each component shown in Figure~\ref{fig:SELECTDiag} and Algorithm~\ref{alg:algorithm} may be implemented in different ways. In this section we discuss the overall algorithm and in the next section we discuss component implementation.

\textbf{State Representation} SELeCT must accumulate experience as states, such that we can identify the conditions where each state is relevant.
SELeCT models each state $S_i$ as a pair, $\langle m_i, \zeta_i \rangle$, an incremental classifier~\cite{gama2003accurate} $m_i$ and concept representation $\zeta_i$. Experience can be accumulated by training $m_i$ on new observations drawn from a given concept, and can be applied by making predictions using $m_i$. 
The \textit{concept representation} $\zeta_i$ defines the distribution each state is relevant to using a similarity function $sim^{\prime}$, \textit{e.g.,} cosine similarity (Eq.~\ref{eq:similarity}). 
We may calculate $sim^{\prime}$ between two representations to estimate the distance between the distributions they represent, and thus estimate relevance.
Different concept representations may capture different information about a distribution, for example, a supervised representation capturing $p(y|X)$ may be used when labels are available, otherwise an unsupervised representation capturing only $p(X)$ may be required~\cite{halstead2021fingerprinting}.

A concept representation $\zeta_i$ for a given state $S_i$ is learned by maximizing similarity to observations seen when $S_i$ is active.
A \textit{background} state $B$ is maintained to represent a sliding window $\omega$ of the most recent observations. We discuss details of the state representation used in our implementation in the next section.
At each time $t$, SELeCT computes the relevance of each state $S_i \in R \cup B$,  to select $S_A^t$ for the next observation using a probabilistic Bayesian approach.

\textbf{State Probability} 
To identify the optimal active state, at each time $t$, SELeCT computes the probability of each state $S_i \in R \cup B$ being optimal at the next time step, $p(S^{t+1}_o = S_i)$, given current knowledge of the system, $K^t$.
General or domain specific knowledge describing concept drift characteristics can be captured in $K^t$, \textit{i.e.,} transition patterns.
Given $K^t$ and the current observation $\langle X^t, y^t \rangle$, the goal is to compute the probability $p(S^{t+1}_o = S_i| K^t, \langle X^t, y^t \rangle)$. 
SELeCT uses a Bayesian approach to break this into two sub-steps, computing the prior probability $p_i$ of $S_i$ being active on the next observation given current knowledge, $p_i = p(S^{t+1}_o = S_i| K^t)$, and computing the likelihood $l_i$ of the current observation being generated from the concept described by $S_i$, $l_i = p(\langle X^t, y^t \rangle | S^{t+1} = S_i)$. 
Bayes Theorem can combine these into a posterior probability:
\begin{equation}\label{eq:bayes}
p(S^{t+1}_o | K^t, \langle X^t, y^t \rangle) \sim p(\langle X^t, y^t \rangle | S^{t+1})p(S^{t+1}_o | K^t)    
\end{equation}

\textbf{Continuous Selection} Finally, SELeCT uses a continuous selection test to select the next active state based on the computed posterior probabilities of each state. A simple \textit{maximum a posteriori} approach, selecting the state with maximum posterior probability, suffers from issues with temporal stability, where noise may cause an erroneous transition to, or a failure to transition away from, a sub-optimal active state. Both cases make it difficult to learn a distinct state for each concept as they disrupt our learning of $\zeta_A$, meaning $S_A$ cannot be accurately recalled in the future.
We instead use a hypothesis test, with the null hypothesis that the current active state had a higher probability over recent observations than any alternative state, giving a guarantee, up to some risk level, that the active state at any given observation is the optimal achievable state for a given $sim^\prime$ from the current set of possible states.

\textbf{Theoretical Analysis: Time and Memory Complexity}
While the time complexity of SELeCT depends on the implementation of each component, we may assume that components follow standard online learning restrictions of constant time and memory complexity per observation.
In this case, SELeCT has the same time complexity as the standard adaptive learning framework.
The active state classification and training steps are $O(1)$ per observation.
Computing state probability can be assumed to be $O(1)$ per observation per stored state if implemented with a fixed amount of memory. Across a state repository of size, $|R|$, SELeCT is then $O(|R|)$ per observation. 
Repository size can be fixed to a constant $|R|$ while retaining performance by implementing standard adaptive learning memory management techniques~\cite{chiu2018diversity, halstead2021recurring}, which we omit for simplicity.  Under these implementation assumptions, the SELeCT framework has a constant time complexity per observation making it suitable for online use.

For comparison, the standard framework replaces the probability calculation with a drift detection step and runs a re-identification step if an alert is triggered. Given a $d$ chance of triggering a drift per observation, the time complexity of the standard framework is $O(d|R|)$ per observation. This is also constant time per observation under the same implementation constraints.
SELeCT has the same constant memory complexity as the standard framework given a fixed size repository, storing $|R|$ states in the repository and one background state, $O(|R| + 1)$.
We empirically validate that the time and memory performance of our implementation of SELeCT is equivalent to existing methods, discussed in the supplementary materials available at https://github.com/BenHals/SELeCT.
\begin{table*}[t]
\caption{Performance against Baselines as Mean$\pm$ Std. The best non-upper bound system in each dataset (row) is bolded.}
\label{tab:baselineExp}
    \centering

\small
\begin{tabular}{llllllllll}
\toprule
     &  &      \multicolumn{1}{c}{FiCSUM} &         \multicolumn{1}{c}{RCD} &         \multicolumn{1}{c}{CPF} &         \multicolumn{1}{c}{DWM} &       \multicolumn{1}{c}{DYNSE} &      \multicolumn{1}{c}{SELeCT} &       \multicolumn{1}{c}{LB} &       \multicolumn{1}{c}{UB} \\
\midrule
\multirow{7}{*}{\rotatebox[origin=c]{90}{$\kappa$}} & AQS &  0.92$\pm$0.04 &  0.72$\pm$0.03 &  0.92$\pm$0.03 &  0.90$\pm$0.01 &  0.79$\pm$0.02 &  \textbf{0.94}$\pm$0.00 &  0.71$\pm$0.03 &  0.95$\pm$0.00 \\
     & AQT &  0.44$\pm$0.06 &  0.39$\pm$0.04 &  0.50$\pm$0.03 &  \textbf{0.57}$\pm$0.01 &  0.47$\pm$0.04 &  \textbf{0.57}$\pm$0.05 &  0.33$\pm$0.02 &  0.61$\pm$0.01 \\
     & AD &  0.85$\pm$0.03 &  0.71$\pm$0.08 &  \textbf{0.87}$\pm$0.03 &  0.85$\pm$0.03 &  0.45$\pm$0.16 &  0.85$\pm$0.02 &  0.82$\pm$0.04 &  0.90$\pm$0.01 \\
     & CMC &  0.22$\pm$0.05 &  0.16$\pm$0.02 &  0.19$\pm$0.03 &  0.22$\pm$0.02 &  0.19$\pm$0.03 &  \textbf{0.25}$\pm$0.03 &  0.24$\pm$0.02 &  0.27$\pm$0.02 \\
     & STGR &  \textbf{0.98}$\pm$0.02 &  0.91$\pm$0.07 &  0.87$\pm$0.04 &  0.92$\pm$0.01 &  0.77$\pm$0.01 &  \textbf{0.98}$\pm$0.00 &  0.93$\pm$0.00 &  0.98$\pm$0.00 \\
     & TREE &  0.35$\pm$0.10 &  0.23$\pm$0.03 &  0.27$\pm$0.05 &  0.30$\pm$0.05 &  0.34$\pm$0.03 &  \textbf{0.50}$\pm$0.06 &  0.25$\pm$0.03 &  0.56$\pm$0.05 \\
     & WIND &  0.90$\pm$0.01 &  0.70$\pm$0.07 &  0.78$\pm$0.02 &  0.86$\pm$0.00 &  0.58$\pm$0.00 &  \textbf{0.92}$\pm$0.01 &  0.89$\pm$0.03 &  0.94$\pm$0.00 \\
\midrule
\multirow{7}{*}{\rotatebox[origin=c]{90}{C-F1}} & AQS &  0.76$\pm$0.10 &  0.29$\pm$0.02 &  0.51$\pm$0.07 &  0.29$\pm$0.00 &  0.29$\pm$0.04 &  \textbf{0.87}$\pm$0.06 &  0.29$\pm$0.00 &  0.97$\pm$0.00 \\
     & AQT &  0.63$\pm$0.10 &  0.26$\pm$0.04 &  0.46$\pm$0.05 &  0.29$\pm$0.00 &  0.22$\pm$0.03 &  \textbf{0.89}$\pm$0.04 &  0.29$\pm$0.00 &  0.97$\pm$0.00 \\
     & AD &  0.82$\pm$0.09 &  0.32$\pm$0.02 &  0.67$\pm$0.09 &  0.29$\pm$0.00 &  0.45$\pm$0.01 &  \textbf{0.91}$\pm$0.03 &  0.29$\pm$0.00 &  0.88$\pm$0.00 \\
     & CMC &  0.73$\pm$0.12 &  0.47$\pm$0.07 &  0.51$\pm$0.06 &  0.67$\pm$0.00 &  0.57$\pm$0.05 &  \textbf{0.84}$\pm$0.07 &  0.67$\pm$0.00 &  0.88$\pm$0.00 \\
     & STGR &  0.95$\pm$0.06 &  0.46$\pm$0.03 &  0.60$\pm$0.12 &  0.50$\pm$0.00 &  0.33$\pm$0.03 &  \textbf{0.98}$\pm$0.02 &  0.50$\pm$0.00 &  0.98$\pm$0.00 \\
     & TREE &  0.59$\pm$0.19 &  0.31$\pm$0.03 &  0.43$\pm$0.03 &  0.29$\pm$0.00 &  0.27$\pm$0.02 &  \textbf{0.94}$\pm$0.06 &  0.29$\pm$0.00 &  0.98$\pm$0.00 \\
     & WIND &  0.98$\pm$0.00 &  0.35$\pm$0.04 &  0.32$\pm$0.05 &  0.40$\pm$0.00 &  0.29$\pm$0.02 &  \textbf{0.99}$\pm$0.01 &  0.40$\pm$0.00 &  0.98$\pm$0.00 \\
\bottomrule
\end{tabular}
\end{table*}

\section{Component Implementation}
In this section, we propose methods for each component of the SELeCT framework. SELeCT requires three main components, a method of representing a concept as a system state, a method of computing state priors and likelihoods, and a continuous selection statistical test. All code is available at https://github.com/BenHals/SELeCT.

\textbf{System State}
We base our state representation on the representation proposed in FiCSUM~\cite{halstead2021fingerprinting}, where each $\zeta_i$ describes the distribution of a concept $C_i$ using a vector of meta-features.
Each \textit{meta-feature} in $\zeta_i$ is the result of a summary function calculated over a set of observations drawn from $C_i$, for example, the mean of $X$ or the variance in $y$, describing one aspect of how a concept behaves over time.
We use the default set of meta-feature functions from FiCSUM to represent a general range of concept behaviours.
We use a weighted cosine distance, with meta-feature weights $W$ calculated as in FiCSUM~\cite{halstead2021fingerprinting}, to calculate the similarity between $\zeta$, approximating the difference between distributions $C_i$ and $C_j$ as
\begin{equation}\label{eq:similarity}
sim^{\prime} (\zeta_i, \zeta_j) = \frac{W\zeta_i \cdot W\zeta_j}{||W\zeta_i|| \cdot ||W\zeta_j||}.    
\end{equation}
 
To train the active representation $\zeta_A$, we monitor a sliding window of recent observations, $\omega$, and a lagged sliding window $B$. The representation $\zeta_B$ built on $B$ is a stable representation of the active concept, as long as we have not detected concept drift since it was captured.
We update the active $\zeta_A$ as the online mean of $\zeta_B$, captured every $|B|$ steps.




\textbf{State Priors}
We calculate the prior probability of a state $S_j$ at time $t+1$, \textit{i.e.,} the probability $p(S_o^{t+1} = S_j | K^{t})$, using a transition matrix based $K^{t}$ which is informed by a concept drift detector.
Given the active state at the current time step, $S_A^{t}$, the probability of $S_j$ being the next optimal state $S_o^{t+1}$ is given by the probability $p(S_o^{t+1} = S_j| S_A^{t}, D^{t})$ of observing a transition from $S_A^{t}$ to $S_j$ in current conditions, \textit{i.e.,} when recent observations are stationary or drifting. We model current conditions as $D^{t}$ using a drift detector, setting $D^{t}$=1 in periods where a drift detector detects significant changes in the likelihood of $S_A^t$, or $D^{t}$=0 otherwise.
We calculate $p(S_o^{t+1} = S_j| S_A^{t}, D^{t})$ using two transition matrices, $TM^1$ and $TM^0$ which capture transitions when $D^t$ is $1$ or $0$ respectively. 
Each entry $TM^d_{ij}$ represents the number of transitions seen from $S_A^{t}$=$S_i$ to $S_j$ when $D^{t}$=$d$. 
We calculate the prior probability for each $S_j \in R$ as 
\begin{equation}
    p(S_o^{t+1} = S_j| S_A^{t} = S_i, D^{t} = d) = \frac{TM^d_{ij}}{\sum_{S_k \in R} TM^d_{ik}}.
\end{equation}


We also use three adjustment parameters. Firstly, a minimum prior probability allows unobserved transitions to occur. Secondly, a `backwards transition' prior of strength $b$ allows a new state to transition back to the previous state to repair false positive transitions, implemented by setting $TM^d_{ij} = b$ when we first observe a transition to state $j$.
Finally, we track transitions of up to $m$ steps, \textit{e.g.,} if transitions A to B and B to C have been observed, we assign some prior probability to the 2-step transition A to C.
Probability for $m$ steps is given by $(TM^d)^m$.
We calculate the transition count as $max(TM^d, (TM^d)^2, \dots, (TM^d)^m)$.
The SELeCT framework enables further information to be incorporated into $K^{t}$, \textit{e.g.,} stream volatility, which has been used to proactively predict concept drift locations~\cite{chen2016proactive}, could be used to update the prior probability of $S_A^t$.

\textbf{State Likelihood}
We calculate relevance as the likelihood of drawing a sliding window of recently observed data, $\omega$, from the concept described by each system state as $p( \omega | S^t = S_j)$. 
We describe $\omega$ by computing the state representation $\zeta_{\omega}$.
A similarity $a_j$ can be calculated between each state representation $\zeta_j$ for $S_j \in R$ and $\zeta_{\omega}$ as $sim^{\prime}(\zeta_j, \zeta_{\omega})$, measuring the similarity between the experience relevant to $S_j$ and the distribution generating incoming data.
We calculate the probability of observing a similarity as high as $a_j$ under the null hypothesis that $S_j$ is relevant. While each state $S_j$ is active, we model the distribution of $a_j$ as a Gaussian, $A_j~\sim N(\mu_j, \sigma_j)$ by capturing the mean and standard deviation of $a_j$ as $\mu_j$ and $\sigma_j$. 
$A_j$ models the distribution of similarity values we would expect to see if $S_j$ was relevant.
We compute the likelihood of drawing a given $a_j$ from $A_j$, to get the likelihood of seeing a $\omega$ with a similarity to $S_j$ of at least $a_j$, if $S_j$ is relevant.
Using Equation~\ref{eq:bayes}, we combine this likelihood with the state prior to get the probability $p(S^{t+1}_o | S^t, \omega)$ that each state is relevant.

\begin{table*}[t]
    \caption{Performance varying drift width, noise and transition noise. Mean $\pm$ Std, best non upper bound method is bolded.}
    \label{tab:variables}
    \centering

\setlength{\tabcolsep}{5pt}
\renewcommand{\arraystretch}{1}\small

\begin{tabular}{lll|rrr|rrr|rrr}
\toprule
          &    \multirow{2}{*}{\rotatebox[origin=c]{90}{Data}}   & & \multicolumn{3}{c}{Drift Width} & \multicolumn{3}{c}{Noise} & \multicolumn{3}{c}{Transition Noise} \\
          &       & Classifer &     \multicolumn{1}{c}{0}    &     \multicolumn{1}{c}{500}  &     \multicolumn{1}{c}{2500} &     \multicolumn{1}{c}{0.00}    &     \multicolumn{1}{c}{0.10}    &     \multicolumn{1}{c}{0.25}    &          \multicolumn{1}{c}{0.00}    &     \multicolumn{1}{c}{0.10}    &     \multicolumn{1}{c}{0.25}    \\
\midrule
\multirow{6}{*}{\rotatebox[origin=c]{90}{$\kappa$}} & \multirow{3}{*}{\rotatebox[origin=c]{90}{TREE}} & Upper &  0.63$\pm$0.04 &  0.62$\pm$0.04 &  0.56$\pm$0.04 &  0.63$\pm$0.04 &  0.53$\pm$0.04 &  0.40$\pm$0.03 &       0.63$\pm$0.04 &  0.63$\pm$0.05 &  0.63$\pm$0.05 \\
          &       & FiCSUM &  0.37$\pm$0.11 &  0.40$\pm$0.11 &  0.26$\pm$0.05 &  0.37$\pm$0.11 &  0.32$\pm$0.10 &  0.25$\pm$0.07 &       0.37$\pm$0.11 &  0.41$\pm$0.11 &  0.41$\pm$0.11 \\
          &       & SELeCT &  \textbf{0.55}$\pm$0.06 &  \textbf{0.53}$\pm$0.07 &  \textbf{0.45}$\pm$0.05 &  \textbf{0.55}$\pm$0.07 &  \textbf{0.45}$\pm$0.08 &  \textbf{0.35}$\pm$0.04 &       \textbf{0.55}$\pm$0.07 &  \textbf{0.55}$\pm$0.06 &  \textbf{0.54}$\pm$0.07 \\
\cmidrule{2-12}
          & \multirow{3}{*}{\rotatebox[origin=c]{90}{AQS}} & Upper &  0.95$\pm$0.00 &  0.95$\pm$0.00 &  0.90$\pm$0.01 &  0.95$\pm$0.00 &  0.84$\pm$0.00 &  0.70$\pm$0.00 &       0.95$\pm$0.00 &  0.95$\pm$0.00 &  0.95$\pm$0.00 \\
          &       & FiCSUM &  0.93$\pm$0.05 &  0.90$\pm$0.05 &  0.76$\pm$0.06 &  0.93$\pm$0.05 &  0.76$\pm$0.08 &  0.59$\pm$0.07 &       0.93$\pm$0.05 &  \textbf{0.94}$\pm$0.02 &  0.93$\pm$0.06 \\
          &       & SELeCT &  \textbf{0.94}$\pm$0.01 &  \textbf{0.93}$\pm$0.01 &  \textbf{0.87}$\pm$0.02 &  \textbf{0.94}$\pm$0.00 &  \textbf{0.82}$\pm$0.05 &  \textbf{0.68}$\pm$0.02 &       \textbf{0.94}$\pm$0.00 &  \textbf{0.94}$\pm$0.01 &  \textbf{0.94}$\pm$0.01 \\
\midrule
\multirow{6}{*}{\rotatebox[origin=c]{90}{C-F1}} & \multirow{3}{*}{\rotatebox[origin=c]{90}{TREE}} & Upper &  0.98$\pm$0.00 &  0.98$\pm$0.00 &  0.98$\pm$0.00 &  0.98$\pm$0.00 &  0.98$\pm$0.00 &  0.98$\pm$0.00 &       0.98$\pm$0.00 &  0.98$\pm$0.00 &  0.98$\pm$0.00 \\
          &       & FiCSUM &  0.59$\pm$0.19 &  0.64$\pm$0.18 &  0.40$\pm$0.11 &  0.59$\pm$0.19 &  0.60$\pm$0.20 &  0.62$\pm$0.19 &       0.59$\pm$0.19 &  0.61$\pm$0.19 &  0.59$\pm$0.17 \\
          &       & SELeCT &  \textbf{0.95}$\pm$0.04 &  \textbf{0.93}$\pm$0.05 &  \textbf{0.87}$\pm$0.04 &  \textbf{0.94}$\pm$0.05 &  \textbf{0.93}$\pm$0.12 &  \textbf{0.94}$\pm$0.04 &       \textbf{0.94}$\pm$0.05 &  \textbf{0.94}$\pm$0.06 &  \textbf{0.93}$\pm$0.08 \\
\cmidrule{2-12}
          & \multirow{3}{*}{\rotatebox[origin=c]{90}{AQS}} & Upper &  0.97$\pm$0.00 &  0.97$\pm$0.00 &  0.97$\pm$0.00 &  0.97$\pm$0.00 &  0.97$\pm$0.00 &  0.97$\pm$0.00 &       0.97$\pm$0.00 &  0.97$\pm$0.00 &  0.97$\pm$0.00 \\
          &       & FiCSUM &  0.78$\pm$0.09 &  0.72$\pm$0.09 &  0.56$\pm$0.09 &  0.78$\pm$0.09 &  0.73$\pm$0.10 &  0.72$\pm$0.10 &       0.78$\pm$0.09 &  0.79$\pm$0.06 &  0.78$\pm$0.08 \\
          &       & SELeCT &  \textbf{0.88}$\pm$0.05 &  \textbf{0.85}$\pm$0.05 &  \textbf{0.74}$\pm$0.05 &  \textbf{0.88}$\pm$0.07 &  \textbf{0.86}$\pm$0.08 &  \textbf{0.86}$\pm$0.05 &       \textbf{0.88}$\pm$0.07 &  \textbf{0.87}$\pm$0.06 &  \textbf{0.88}$\pm$0.05 \\
\bottomrule
\end{tabular}


\end{table*}
\textbf{Continuous Selection}
To select the the active state for an incoming observation given $p(S^{t+1}_o | S^t, \omega)$, we compare the probability of $S_A^t$ against $B$ and all states $S_i \in R$. We use a statistical test to guarantee that the selected $S_A^{t+1}$ has the highest probability over a window of recent timesteps, in order to maintain temporal stability even in noisy conditions.
We test a window of the $w_0$ most recent posterior probabilities for the active state, which has mean $\mu_0$, against a window of the $w_1$ most recent probabilities for each alternative state, which has mean $\mu_1$.
Similarly to the ADWIN~\cite{bifet2007learning} concept drift detector, we select each window as the set of recent probabilities with no significant change in mean.
The difference in means between the active and alternative states, $\mu_1 - \mu_0$, is tested against a threshold $\epsilon$ to test whether the alternative is significantly more relevant than the current active state.
We use the Hoeffding bound to select $\epsilon$ based on the size of $w_0$ and $w_1$ in order to constrain the false positive and negative error rate for this test to be a parameter $\delta$~\cite{bifet2007learning}.
We transition to the alternative state with the highest $\mu_1$ at least $\epsilon$ above $\mu_0$.
Any conflict between alternative states is resolved in the next evaluation.
If we transition to $B$, it is added to the repository, and a new $B$ is initialized to represent $\omega$.

\textbf{State Merge} Over time, the repository can accumulate duplicate states representing the same concept~\cite{ahmadi2018modeling}. We remove duplicates by calculating correlations between state probabilities and merge states with a correlation above a threshold parameter of 0.95 by combining entries in the transition matrix and removing the state with less training from the repository.

\section{Evaluation}\label{sec:Evaluation}
Our hypothesis is that by selecting more relevant active states, SELeCT is able to more accurately accumulate and apply experience matching the underlying concept of a stream, enabling increased accuracy.
We evaluate this hypothesis in this section, comparing SELeCT against an optimal reference and alternative streaming methods, and studying the effect of each of SELeCT's components.

\textbf{Evaluation measures}
We evaluate two aspects of learning in changing conditions: classification accuracy using the kappa statistic, $\kappa$, and context tracking performance using the C-F1 measure~\cite{halstead2021fingerprinting}.
C-F1 measures the relevance of the experience chosen by a system to the ground truth concept at each point in a stream, encompassing standard measures like drift detection delay, false alarms, as well as re-identification performance.
For each state $s$ and concept $c$, we calculate the recall and precision of timesteps when $c$ is active against the timesteps where $s$ is active, measuring how much of the experience accumulated by $s$ describes a $c$, and how often did the system identify $s$ as relevant when $c$ was present.
C-F1 ranges from 0 to 1, reporting the average F1 score of the best matching for each $c$ as an overall measure of how well the experience captured by a system matched underlying concepts.

\textbf{Datasets}
We require datasets with known ground truth concepts to measure the use of relevant experience.
We use four real-world datasets~\cite{moreira2018classifying} with known concepts.
The \textit{Aedes} (AQS, AQT) datasets classify insect characteristics, with six different temperature ranges as concepts.
\textit{Arabic-Digets} (AD) classifies a speaker based on speech patterns, with six different digits as concepts.
CMC classifies a survey result, with two age ranges as concepts.
We use three synthetic datasets to investigate specific settings where SELeCT is beneficial, STAGGER (STGR) and RandomTree (TREE) are benchmarks from previous literature, and we provide a new synthetic data generator, WIND, based on a simulated air quality prediction task.
For all datasets we construct streaming datasets by repeating each concept in the dataset three times in an order generated using an underlying transition pattern, which is varied as described in the following experiments, to explore different scenarios. 
We evaluate abrupt and gradual concept drift, and report the mean$\pm$standard deviation over 45 seeds.

\textbf{Experiment Setup}
We implement SELeCT using Hoeffding Tree~\cite{gama2003accurate} and ADWIN~\cite{bifet2007learning} as the base classifier and drift detector, using default parameters in Scikit-Multiflow~\cite{skmultiflow}. We also study a neural network base classifier, implemented in PyTorch.
We use the default state representation proposed in FiCSUM, making FiCSUM a comparable standard framework baseline. 
Parameters for SELeCT were chosen using a linear sensitivity analysis on the \textit{AD} and \textit{CMC} datasets.
 Datasets and parameters are described in supplementary material.

\subsection{Performance evaluation}
We first evaluate SELeCT against existing methods. Table~\ref{tab:baselineExp} shows the $\kappa$ and C-F1 of SELeCT against FiCSUM, CPF, DWM and DYNSE described in Section~\ref{sec:related-work}. We discuss the selection of baselines in the supplementary.  
\textit{UB} reports the performance of a Hoeffding Tree classifier with perfect re-identification and drift detection with a fixed delay of 100 observations to represent the performance upper bound, \textit{i.e.}, selecting the optimal state for each observation. \textit{LB} reports the performance of a non-adaptive Hoeffding Tree classifier to represent the lower bound of adaptive learning performance.

SELeCT achieves a C-F1 at least 80\% of the upper bound. In a Nemenyi significance test, shown in the supplementary material, SELeCT C-F1 performance is not significantly lower than the upper bound, and outperforms all competitors. 
SELeCT achieves a C-F1 of 0.94 in TREE, 63\% higher than FiCSUM's 0.59.
This result validates our hypothesis that SELeCT can select the optimal achievable state. SEleCT achieves a higher $\kappa$ value than all baselines, except CPF in AD. 
SELeCT achieves a $\kappa$ of 0.50 in TREE, 43\% higher than FiCSUM's 0.35, indicating that selecting the optimal estimated state does provide increased classification performance. We observe that higher C-F1 does not always translate into significant accuracy improvements, which we explore in the next experiment.

\begin{figure}[]
    \centering
    \includegraphics[width=0.425\textwidth]{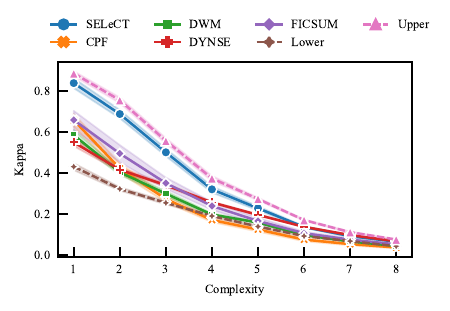}
    \caption{Performance at increasing complexity.}
    \label{fig:complexity}
\end{figure}

\textbf{Concept Complexity Experiment}
In some datasets, recalling relevant experience does not substantially increase accuracy compared to collecting new experience, \textit{e.g.,} if concepts are simple to learn.
We hypothesize that the better active state selection shown by SELeCT may not translate into significantly higher classification performance in these cases.
Evidence for this comes from the small difference between \textit{LB} and \textit{UB} in some datasets, indicating that even perfect adaptive learning does not improve classification performance.
This experiment investigates this effect, reporting the performance of SELeCT against baselines under increasing concept \textit{complexity}.
Using the TREE dataset, where $p(y|X)$ for each concept is given by a randomly generated decision tree, we can increase the depth of these trees to create more complex distributions which take longer to learn.
Figure~\ref{fig:complexity} shows that as complexity increases, the classification performance of all methods falls. However, SELeCT retains a performance closer to \textit{UB}, especially when adaptive learning is more beneficial, \textit{i.e.,} there is a larger gap between \textit{UB} and \textit{LB}.

\textbf{Gradual Drift and Noise Experiments}
In this experiment, we investigate gradual drift and noise, two cases where we hypothesize the standard framework may fail. We test against FiCSUM as a baseline.
Under gradual drift, concept drift occurs over a long period which may be missed by drift detection in the standard framework.
Under noise, drift detection may also be less sensitive, as changes due to drift are obscured by changes due to noise.
Table~\ref{tab:variables} reports performance at increasing concept drift width and with increasing amounts of uniform class noise, on two representative datasets. Both scenarios show a failure case for the standard framework, with the performance of FiCSUM dropping significantly, while SELeCT succeeds in maintaining performance close to the upper bound.

\subsection{Component Evaluation}
 Table~\ref{tab:baselineExp} shows an ablation study comparing the reference SELeCT implementation against variants.

\textbf{Prior Probability}
Our prior probability component assumes that concept transitions follow an underlying transition pattern. We evaluate robustness to this assumption in Table~\ref{tab:variables}, which shows performance with increasing noise levels in underlying concept transitions. SELeCT shows no substantial changes in performance relative to the upper bound.
These results indicate that our prior probability component is robust to the true underlying distribution of concepts.
Table~\ref{tab:ablation} shows that a variant of SELeCT, $S_p$, using an uninformative, uniform prior, is outperformed by our implementation, indicating our prior component is beneficial.

\textbf{Continuous Selection}
To test the effect of continuous selection, Table~\ref{tab:ablation} shows the performance of a variant of SELeCT, $S_{MAP}$, using a n\"{a}ive MAP selection procedure which selects the most probable state at each step rather than using our Hoeffding bound test. SELeCT outperforms this variant, indicating our continuous selection component is beneficial and validating our hypothesis that using the Hoeffding bound to select active states can improve performance. Crucially, we observe a significantly lower C-F1, indicating temporal instability in the selection of active states.

\textbf{State Merging} Table~\ref{tab:ablation} shows that a variant of SELeCT, $S_{m}$, which does not merge states achieves lower performance, especially in C-F1, indicating that by removing redundant representations we can better model underlying context.

\begin{table*}[t]
\caption{Ablation Study, reporting mean$\pm$ std. Best non upper-bound method bolded for each set of runs.}
\label{tab:ablation}
    \centering
\setlength{\tabcolsep}{4pt}
\renewcommand{\arraystretch}{0.8}\small

\begin{tabular}{llrrrr|rrrrrr}
\toprule
     & & \multicolumn{4}{c|}{45 Runs} & \multicolumn{6}{c}{20 Runs} \\
     &  &      \multicolumn{1}{c}{SELeCT} &       \multicolumn{1}{c}{$S_p$} &   \multicolumn{1}{c}{$S_{MAP}$}  &     \multicolumn{1}{c}{$S_{m}$} &     \multicolumn{1}{|c}{SELeCT} &   \multicolumn{1}{c}{$S_{cndpm}^{*}$} & \multicolumn{1}{c}{CNDPM$_{np}^{*}$} & \multicolumn{1}{c}{CNDPM$_{p}^{*}$}  &         \multicolumn{1}{c}{UB} &  \multicolumn{1}{c}{UB$_{cndpm}^{*}$}\\
\midrule
\multirow{7}{*}{\rotatebox[origin=c]{90}{$\kappa$}} & AQS &  \textbf{0.94}$\pm$0.00 &  \textbf{0.94}$\pm$0.00 &  0.91$\pm$0.06 &  \textbf{0.94}$\pm$0.00 &  \textbf{0.94}$\pm$0.00 &  0.66$\pm$0.15 &    0.20$\pm$0.20 &   0.11$\pm$0.07 &  0.95$\pm$0.00 &  0.91$\pm$0.01 \\
     & AQT &  \textbf{0.57}$\pm$0.05 &  0.55$\pm$0.06 &  0.44$\pm$0.11 &  0.56$\pm$0.04 &  \textbf{0.53}$\pm$0.07 &  0.47$\pm$0.08 &    0.30$\pm$0.23 &   0.03$\pm$0.01 &  0.61$\pm$0.01 &  0.60$\pm$0.00 \\
     & AD &  \textbf{0.85}$\pm$0.02 &  \textbf{0.85}$\pm$0.02 &  0.81$\pm$0.05 &  \textbf{0.85}$\pm$0.02 &  \textbf{0.86}$\pm$0.01 &  0.68$\pm$0.11 &    0.57$\pm$0.12 &   0.56$\pm$0.18 &  0.90$\pm$0.01 &  0.89$\pm$0.02 \\
     & CMC &  \textbf{0.25}$\pm$0.03 &  0.23$\pm$0.04 &  0.23$\pm$0.05 &  \textbf{0.25}$\pm$0.03 &  \textbf{0.24}$\pm$0.03 &  0.17$\pm$0.05 &    0.04$\pm$0.03 &   0.10$\pm$0.04 &  0.27$\pm$0.02 &  0.26$\pm$0.02 \\
     & STGR &  \textbf{0.98}$\pm$0.00 &  \textbf{0.98}$\pm$0.01 &  0.92$\pm$0.08 &  \textbf{0.98}$\pm$0.00 &  \textbf{0.98}$\pm$0.00 &  0.87$\pm$0.09 &    0.13$\pm$0.07 &   0.47$\pm$0.12 &  0.98$\pm$0.00 &  0.95$\pm$0.00 \\
     & TREE &  \textbf{0.50}$\pm$0.06 &  0.49$\pm$0.06 &  0.41$\pm$0.11 &  \textbf{0.50}$\pm$0.06 &  \textbf{0.51}$\pm$0.06 &  0.31$\pm$0.05 &    0.23$\pm$0.07 &   0.10$\pm$0.02 &  0.58$\pm$0.03 &  0.38$\pm$0.02 \\
     & WIND &  \textbf{0.92}$\pm$0.01 &  \textbf{0.92}$\pm$0.01 &  0.89$\pm$0.03 &  \textbf{0.92}$\pm$0.01 &  \textbf{0.92}$\pm$0.01 & 0.54$\pm$0.05 &   -0.00$\pm$0.00 &   0.00$\pm$0.00 &  0.94$\pm$0.00 &  0.62$\pm$0.00 \\
\midrule
\multirow{7}{*}{\rotatebox[origin=c]{90}{C-F1}} & AQS &  \textbf{0.87}$\pm$0.06 &  \textbf{0.87}$\pm$0.04 &  0.79$\pm$0.10 &  0.86$\pm$0.06 &  \textbf{0.85}$\pm$0.06 &  0.77$\pm$0.11 &    0.28$\pm$0.00 &   0.29$\pm$0.00 &  0.97$\pm$0.00 &  0.97$\pm$0.00 \\
     & AQT &  \textbf{0.89}$\pm$0.04 &  0.87$\pm$0.07 &  0.71$\pm$0.12 &  0.88$\pm$0.04 &  0.88$\pm$0.03 &  \textbf{0.89}$\pm$0.07 &    0.28$\pm$0.01 &   0.29$\pm$0.00 &  0.97$\pm$0.00 &  0.97$\pm$0.00\\
     & AD &  \textbf{0.91}$\pm$0.03 &  0.89$\pm$0.04 &  0.85$\pm$0.07 &  0.88$\pm$0.03 &  \textbf{0.91}$\pm$0.01 &  0.88$\pm$0.03 &    0.28$\pm$0.01 &   0.29$\pm$0.00 &  0.88$\pm$0.00 &  0.88$\pm$0.00 \\
     & CMC &  \textbf{0.84}$\pm$0.07 &  0.76$\pm$0.11 &  0.82$\pm$0.09 &  0.80$\pm$0.07 &  0.83$\pm$0.08 &  \textbf{0.84}$\pm$0.08 &    0.62$\pm$0.01 &   0.67$\pm$0.00 &  0.88$\pm$0.00 &  0.88$\pm$0.00\\
     & STGR &  \textbf{0.98}$\pm$0.02 &  \textbf{0.98}$\pm$0.02 &  0.87$\pm$0.13 &  0.97$\pm$0.02 &  \textbf{0.98}$\pm$0.01 &  0.92$\pm$0.09 &    0.37$\pm$0.03 &   0.51$\pm$0.03 &  0.98$\pm$0.00 &  0.98$\pm$0.00\\
     & TREE &  \textbf{0.94}$\pm$0.06 &  \textbf{0.94}$\pm$0.06 &  0.78$\pm$0.17 &  0.93$\pm$0.05 &  0.94$\pm$0.07 &  \textbf{0.96}$\pm$0.02 &    0.26$\pm$0.01 &   0.29$\pm$0.00 &  0.98$\pm$0.00 &  0.98$\pm$0.00\\
     & WIND &  \textbf{0.99}$\pm$0.01 &  0.98$\pm$0.02 &  0.95$\pm$0.07 &  0.97$\pm$0.01 &  \textbf{0.99}$\pm$0.01 &  \textbf{0.99}$\pm$0.01 &    0.40$\pm$0.00 &   0.40$\pm$0.00 &  0.98$\pm$0.00 &  0.98$\pm$0.00\\
\bottomrule
\end{tabular}
\end{table*}

\subsection{Continual Learning Comparison}
We evaluate SELeCT against CNDPM~\cite{lee2020neural}, an expansion based neural network continual learning method able to allocate new parameters to accumulate experience from new tasks. CNDPM can identify which task is relevant to each observation, but does not consider task relevance to change over time, \textit{i.e.,} cannot forget irrelevant tasks or recall different tasks for the same input. CNDPM includes a prior probability based on the number of observations drawn from each task. In a streaming setting, we find this prior undervalues new concepts relative to existing concepts, making them temporally unstable and difficult to learn. We study two variants of CNDPM, with and without the prior, as CNDPM$_{p}$ and CNDPM$_{np}$. To control for the effect of using a neural network rather than a Hoeffding tree, we study variants of SELeCT and the upper bound classifier, denoted $S_{cndpm}$ and UB$_{cndpm}$ respectively, which use the CNDPM base neural network to learn each state. We use the same default hyper-parameters for all networks. Due to the reduced efficiency of the neural network compared to the base Hoeffding tree, we performed 20 runs of each experiment rather than 45. Table~\ref{tab:ablation} reports the mean and standard deviation. 

We observe that in both SELeCT and UB the neural network base classifier reduces performance, however, in both cases, an adaptive learning approach is still able to learn, highlighting the generality of our framework. We observe that removing the prior increases the performance of CNDPM in most cases, however in all cases performs significantly worse than SELeCT, even using the same base classifier. This result verifies our hypothesis that existing methods cannot adequately handle changes and recurrences in the relevance of experience over time, leading to degraded performance in a streaming environment. In contrast, SELeCT is, in many cases, within one standard deviation of an upper-bound system. 

\section{Related Work}~\label{sec:related-work}
Our problem formulation shares similarities to dynamic classifier selection~\cite{cruz2018dynamic, de2016handling} and continual learning~\cite{parisi2019continual}.
Dynamic selection considers the joint distribution of data to be partitioned on $X$ into \textit{regions of competence} specialized in handling distinct inputs~\cite{almeida2018adapting, tsymbal2008dynamic}, \textit{i.e.,} given a region, $X$, relevant experience describes $p(y|X)$.
Here context $H^t$ determines the region data is drawn from, $p(X|H^t)$, but the relevance of experience within a region, $p(y|X)$, is constant, \textit{i.e.,} forgetting is not required as real concept drift~\cite{gama2014survey} is not considered.
DYNSE~\cite{de2016handling} considers concept drift, estimating recent experience as more relevant to handle gradual change in $p(y|X)$ over time, but not recurrences. 

The aim of Continual learning is to learn new tasks without forgetting information relevant to past tasks~\cite{parisi2019continual}. A major research focus is avoiding catastrophic forgetting.
Many continual learning approaches are not practical for streaming tasks as they assume that a task ID encoding relevant experience is known for each observation. 
Recent \textit{task-free} continual learning approaches~\cite{lee2020neural} can identify experience relevant to a task.
However, similarly to dynamic selection, continual learning determines the relevance of past experience to $X$, rather than to the current concept. While the distributions of observations, $p(X)$, is considered to change over time as observations arrive from different tasks, each specific observation is associated with the same task over the entire stream, \textit{i.e.,} $p(y|X)$ is constant. In contrast, under real concept drift $p(y|X)$ changes over time, \textit{i.e.}, a particular observation is associated with different concepts at different points in time and must be predicted using different experience.
 Some recent methods update experience over time~\cite{lee2020neural, de2021continual}, implicitly forgetting experience to adapt to gradual concept drift. However, no existing continual learning methods can explicitly identify irrelevant experience or store and recall relevant experience to learn recurring concepts. 
In Section~\ref{sec:Evaluation}, we find that current continual learning methods struggle to learn in a streaming setting with recurring concepts. We note that our setting is different from modeling recurrent change points~\cite{maslov2017blpa}.

The adaptive learning framework described in Section~\ref{sec:background} has been used in many approaches, usually with new methods of representing the active state and monitoring changes in its relevance. RCD~\cite{gonccalves2013rcd}, uses an accuracy based similarity measure to monitor the relevance of the active state continuously, while a distribution similarity test is used during re-identification to evaluate the relevance of stored states.
JIT~\cite{alippi2013just} and CPF~\cite{anderson2016cpf} monitor the active state using error rate and feature distribution, while relevance for re-identification is based on hypothesis testing and classifier equivalence~\cite{gama2014recurrent}. GraphPool~\cite{ahmadi2018modeling} additionally uses a concept transition matrix when evaluating the relevance of stored states.
\cite{wang2006improving} and \cite{angel2016predicting} use a similar transition matrix approach to identify recurring concepts.
FiCSUM~\cite{halstead2021fingerprinting} uses a vector similarity measure to monitor the active state continuously and sparsely evaluate stored states.
\cite{gama2014recurrent} propose using a secondary meta-learning classifier trained to predict state relevance. DWM~\cite{kolter2007dynamic} and ARF~\cite{gomes2017adaptive} use an ensemble active state, using error rate to monitor the relevance of each member ~\cite{de2019overview, krawczyk2017ensemble}, but have no mechanism to recall forgotten experience once a given classifier is dropped from the ensemble.

\section{Conclusion}
Adaptive learning provides a framework for data stream classification in the presence of concept drift by identifying and reusing previous experience relevant to current conditions. However, existing methods encounter common failure cases where sub-optimal experience is selected due to sparse and binary evaluation, hindering our ability to learn from streaming data in changing conditions.
We propose SELeCT, a probabilistic framework that is able to avoid these failure cases by continuously evaluating the relevance of all states to select the optimal state for each observation.
Our evaluation shows the relevance of the experience chosen by SELeCT  to ground truth concepts is comparable to a perfect knowledge baseline, and up to 60\% higher than five baseline methods, enabling us to learn in new scenarios featuring complex changing and recurring conditions.
Experiments varying data complexity, noise, and drift width show more accurate identification of relevant experience allows SELeCT to achieve a $\kappa$ statistic up to 43\% higher than alternative systems.

\section*{Acknowledgment}
The work was supported by the Marsden Fund Council from New Zealand Government funding (Project ID 18-UOA-005), managed by Royal Society Te Ap{\=a}rangi.

\bibliographystyle{IEEEtran}
\bibliography{references.bib}

\end{document}


\maketitle

\begin{table}[tb]
    \centering
    \caption{Dataset Characteristics}
    \small
    \begin{tabular}{lrrrr}
    \toprule
    \multicolumn{1}{c}{Dataset} & \multicolumn{1}{c}{Context size} & \multicolumn{1}{c}{\# features} & \multicolumn{1}{c}{\# contexts} & \multicolumn{1}{c}{Dataset Length}\\
    \midrule
    AQT & 4000 & 25 & 6 & 72000\\
    AQS & 4000 & 25 & 6 & 72000\\
    AD & 880 & 10 & 10 & 15840\\
    CMC & 736 & 8 & 2 & 4416\\
    \midrule
    TREE & 5000 & 10 & 6 & 90000\\
    STGR & 5000 & 3 & 3 & 90000\\
    WIND & 5000 & 16 & 6 & 90000\\
    \midrule
    \end{tabular}
    \label{tab:datasets}

\end{table}
\section{Datasets}\label{sec:A-Datasets}
We use four real-world datasets with identified contexts from \cite{moreira2018classifying}.
\begin{itemize}
    \item \textit{Aedes-Culex} - AQT: Contains 25 features measuring the flight patterns of male and female mosquitoes, with a class label describing two different species. Context is given by six temperature ranges, modified in an experimental setting to form six concepts to change flight patterns~\cite{dos2018unsupervised}.
    \item \textit{Aedes-Culex} - AQS: Same as AQT using the sex of mosquito as a class label
    \item \textit{Arabic Digits} (AD): Contains 10 features measuring speech patterns of speakers saying Arabic digits aloud, with speaker gender as class labels. Contexts are the digit being spoken.
    \item \textit{CMC}: Contains 8 features measuring responses to a survey on contraceptive use in Indonesia, with a binary yes or no answer. Two age ranges designate contexts.
\end{itemize}

We use three synthetic datasets, code to recreate each is provided.
\begin{itemize}
    \item \textit{Stagger} (STGR)~\cite{gama2004learning}: Implemented in Scikit-Multiflow~\cite{skmultiflow}, simulates objects with three categorical features, each with an accept or reject label. Concepts are three labeling functions which define acceptance. Concepts may conflict.
    \item \textit{RandomTree} (TREE)~\cite{halstead2021fingerprinting}: 
    Implemented in FiCSUM~\cite{halstead2021fingerprinting},
    generates samples from a decision tree. Concepts use distinct trees, leading to different join probabilities. Sampling is done using a $p(X)$ with a random mean, variance, skew and kurtosis for each concept. Labels are balanced. 
   We define a \textit{complexity} measure based on tree height. A $complexity=d$ has a min height of $d$, and max height of $d+2$. Increasing $d$ generates deeper trees with more complex $p(X, y)$, increasing the number of observations required to learn. We use $d=3$, except for the experiment where complexity was varied.
    Code for our implementation is available in supplementary material.
    \item \textit{WIND}: We propose a new air quality classification data generator able to generate synthetic data with concept drift in both $p(X)$ and $p(y|X)$. WIND simulates a circular set of sensors with a central target sensor. Labels are the quantized air quality of the target, features are current and previous readings from surrounding sensors. Spatio-temporal dynamics of pollution generation and transmission are simulated differently for each concept by placing $n$ pollution sources and setting wind speed and direction which dictates pollution movement and dispersal. Each concept specifies $n$, their placement (upwind of the target), the strength of generation, variance in generation, rate of generation, and a wind speed and direction. Changes in concept can be abrupt, \textit{i.e.,} a new factory has turned on, or gradual, a slow change in wind direction. Code for this synthetic data generator is available open source.
\end{itemize}

Streams are generated by combing observations drawn from each concept in each data source in varying configurations.
We repeat each of the $|C|$ concepts, capped at 6, $r=3$ times for a maximum of 18 concept drifts and 90,000 observations.
The value 3 was chosen as it is large enough to capture recurring concepts in each stream, but small enough that the runtime of each test is not too large, allowing many tests to be run efficiently. We test SELeCTs performance against $r$, finding that SELeCTs performance improves as $r$ increases. 
The order of concepts was based on a random transition matrix based on a circular order to ensure that all concepts are reachable from all other concepts. 
For each concept $c_i$, a transition probability of $p^f$ was assigned to each of the $f \in [1, 2, \dots, F]$ shuffled concepts following $c_i$, with $p$ a probability decay parameter and $F$ the number of forward connections. A transition noise parameter $tn$ introduces random transitions to any concept. We set $p=0.7$, $f=3$ and $tn=0$ for most experiments, to create reasonably complex transition patterns. We evaluate sensitivity to transition noise to measure robustness to the transition pattern.

\section{Baselines}\label{sec:A-Baselines}
We test against FiCSUM, a recent state-of-the-art method using the previous standard adaptive learning framework, and two alternatives: RCD, CPF.
We test DYNSE~\cite{almeida2018adapting} as a dynamic classification method, DWM~\cite{kolter2007dynamic} as an ensemble method, and CNDPM as a recent state-of-the-art task-free continual learning method. Finally, we compare to LB and UB as theoretically minimal and optimal adaptive learning baselines.
\begin{itemize}
    \item RCD~\cite{gonccalves2013rcd}: Default implementation in MOA~\cite{DBLP:journals/jmlr/BifetHKP10} (Java). Default parameters: KNN statistical test for relevance, Hoeffding Tree base classifier (for comparison to SELeCT), default EDDM drift detector (required).
    \item CPF~\cite{anderson2016cpf}: Default implementation in MOA (Java). Default parameters: Hoeffding Tree base classifier.
    \item FiCSUM~\cite{halstead2021fingerprinting}: Default implementation in Scikit-Multiflow~\cite{skmultiflow} (Python). We use the same values for shared parameters for the FiCSUM baseline and SELeCT. Default parameters: Hoeffding Tree base classifier, ADWIN drift detector, default meta-information measures (without IMF, MI and PACF as suggested by authors), default weighted cosine distance similarity measure, default Fisher score feature weighting.
    \item DYNSE~\cite{almeida2018adapting}: Default implementation in MOA (Java). Default parameters from the paper: default KNORA Eliminate strategy. For C-F1 we consider the active state to be the state selected for each batch.
    \item DWM~\cite{kolter2007dynamic}: Default implementation in scikit-multiflow (Python). Default parameters. For C-F1 we consider the active state to be static to represent the single ensemble.
    \item CNDPM~\cite{lee2020neural}: Default implementaion in Pytorch (Python). Default parameters: MLP architecture with 1/4 sized layers to better learn the smaller number of features in our datasets (shared with SELeCT and UB variants). Active state is considered the identified task. We compare to a variant with prior set to 1 for all tasks.
    \item LB: A single scikit-multiflow Hoeffding Tree is run with no adaptation, representing the lower bound of adaptive learning performance.
    \item UB: System with a scikit-multiflow Hoeffding Tree base classifier and perfect information on ground truth concepts with a static 100 observation delay (no delay is not possible in any real system, so this was chosen as a realistic setting). 
    Adapts to recurring concepts by reusing the last classifier and new concepts by initializing a new Hoeffding tree. Achieves the upper bound for adaptive learning given the Hoeffding Tree base classifier and a 100 delay.
\end{itemize}

\section{Experiment setup}\label{sec:A-experiment}
All experiments were repeated 45 times, using the numeric seeds 1 to 45 to seed the Numpy random number generator, using Numpy version 1.20. All systems are evaluated on the same datasets.

\textbf{Environment}
All experiments were conducted on an AMD Ryzen Threadripper PRO 3995WX CPU with 64 cores running at 2.70 GHz, with 256 Gb of RAM.
Each process was assigned to a single virtual core to ensure consistent running time. Experiments were run using Python version 3.7 or Java version 16. The Scikit-Multiflow framework version used was 0.5.3, and the MOA framework version was 2020.07.1.

\textbf{Source Code}
All source code is available at https://bit.ly/3gqSKL5. This includes source code for SELeCT, as well as evaluation and dataset creation scripts, including the WIND data generator. All instructions to run SELeCT as well as demo scripts recreating results and visualizing the proposed WIND dataset are given.

\textbf{Time \& Memory} We evaluate the time and memory of each system, finding SELeCT has equivalent runtime and half the memory use of FiCSUM, taking an average over all datasets of 1406s and 96MB compared to FiCSUMs 1452s and 176MB. Full results are shown in the source code repository.

\section{Parameter Selection}\label{sec:A-Parameters}
Our SELeCT implementation inherits parameters from the FiCSUM concept representation $\zeta$ and ADWIN drift detection, and also requires new parameters, as described in the main paper. 
We do not see substantial differences in optimal parameters across the datasets tested (CMC and AD), thus propose a set of general default parameters.
We use these parameters for all experiments in the paper, and do not tune them to specific datasets.

We selected default parameters using linear searches on the CMC dataset, validated on the AD dataset. For each parameter a valid range was selected and 20 evenly spaced values were tested for accuracy and C-F1 on CMC. The value with the highest harmonic mean of accuracy and C-F1 was selected. Table~\ref{tab:parameters} lists all parameters in our implementation, values tested and final value. The risk parameters describe the risk level for the Hoeffding bound test and window selection in the continuous selection method, and the state grace period is a number of observations to pause selection when a new state in initialized to allow its representation to stabilize. Min state likelihood is a small value to avoid 0 value likelihoods collapsing probabilities. Prior parameters were described in the implementation. The merge correlation sets the threshold before states are merged, we set this to a high value to keep merging conservative, to avoid merging representations of different concepts.
\begin{table}
    \centering
    \caption{SELeCT implementation parameter values}
    \small
    \begin{tabular}{lrrr}
    \toprule
        Parameter & Component & Range & Value  \\
    \midrule
        \multicolumn{4}{c}{Outside Components} \\
    \midrule
        Sensitivity & ADWIN & Default & 0.05 \\
        Window size  &  $\zeta$ & Default & 100 \\
        Buffer Ratio   & $\zeta$ & Default & 0.2 \\
        \midrule
            \multicolumn{4}{c}{Our Components} \\
        \midrule
        Hoeffding bound risk   & Selection &0.05 - 0.9 & 0.75 \\
        Min state likelihood  & Likelihood & 0.0001 - 0.01 & 0.005 \\
        $B$ prior multiplier  & Prior & 0.05 - 1.0 & 0.4 \\
        Min prior  & Prior & 0.05 - 1.0 & 0.7 \\
        Multihop prior multiplier  & Prior &  0.05 - 1.0 & 0.7 \\
        Prev state prior  & Prior & 0 -500 & 50 \\
        Merge correlation  & Merge & 0.05 - 1.0 & 0.95 \\
    \bottomrule
    \end{tabular}
    
    \label{tab:parameters}
\end{table}

\section{Significance Testing}
\begin{figure}[tbh]
    \centering
    \includegraphics[width=0.4\textwidth]{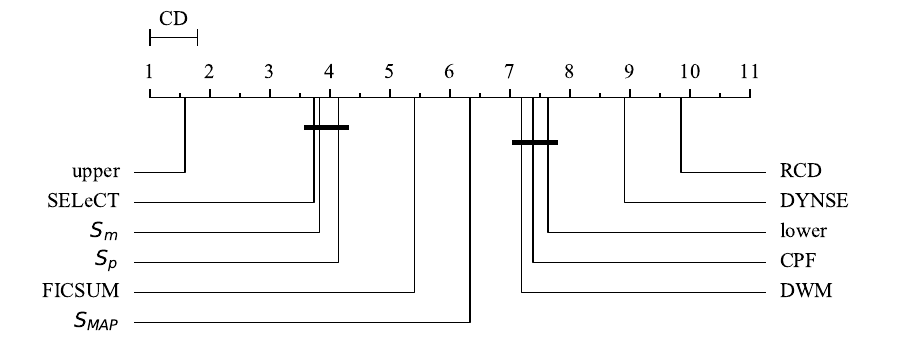}
    \caption{Critical difference: $\kappa$-statistic across systems}
    \label{fig:kappasig}
\end{figure}
\begin{figure}[tbh]
    \centering
    \includegraphics[width=0.4\textwidth]{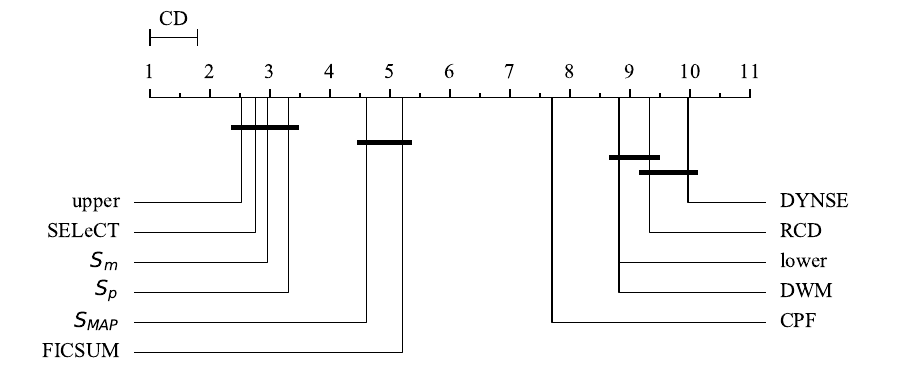}
    \caption{Critical difference: C-F1 across systems}
    \label{fig:cf1sig}
\end{figure}

\bibliographystyle{IEEEtran}
\bibliography{references.bib}